\documentclass[sigconf]{acmart}
\usepackage{multirow}
\setcopyright{none}
\AtBeginDocument{%
  }

\begin{document}

\title{Temporary Adaptation with Partial Feedback for Long-Term Value Prediction in Live Streaming}

\author{Chen Zhang}
\author{Weihao Liu}
\author{Xiaopeng Ye}
\affiliation{%
  \department{\mbox{Gaoling School of Artificial Intelligence}}
  \institution{Renmin University of China}
  \city{Beijing}
  \country{China}
}
\email{\{zhangchen111,weihaoliu,xpye\}@ruc.edu.cn}

\author{Haiyuan Zhao}
\affiliation{%
  \institution{Independent Researcher}
  \city{Beijing}
  \country{China}
}
\email{haiyuanzhao\_suke@163.com}

\author{Xiao Zhang}
\authornote{Xiao Zhang is the corresponding author (e-mail: zhangx89@ruc.edu.cn). 
}
\author{Jun Xu}
\affiliation{%
 \department{\mbox{Gaoling School of Artificial Intelligence}}
  \institution{Renmin University of China}
  \city{Beijing}
  \country{China}
}
\email{\{zhangx89,junxu\}@ruc.edu.cn}

\begin{abstract}

In live-streaming recommendation, users repeatedly interact with authors, forming lasting relationships.
Predicting the long-term value (LTV) of these user--author pairs is crucial for sustaining engagement and supporting platform growth.
However, LTV labels become mature only after a fixed observation window.
Models trained only on mature labels face a supervision lag: they serve current requests using supervision from older interactions, limiting their ability to adapt to recent distribution shifts.
Recent immature samples provide partial feedback that can help reduce this lag.
Existing methods either jointly model partial feedback and LTV targets or generate pseudo LTV labels for training.
However, joint modeling may introduce conflicting gradients between objectives, while retaining pseudo-label updates can cause estimation errors to accumulate.
 Since the corresponding full labels eventually become available,
partial feedback can be used for temporary adaptation to mitigate
supervision lag, while only parameter updates based on mature
true labels are retained for subsequent training.

We propose \textbf{DELTA}, a framework that uses partial feedback to generate pseudo LTV labels for adaptation on recent samples and decouples this temporary adaptation from persistent learning on mature true labels.
Specifically, we train a privileged pseudo-labeler (PPL) that uses partial feedback as privileged information to generate pseudo LTV labels for recent samples, providing supervision for adaptation.
DELTA maintains a base model trained only on mature
true labels. Before each serving period, DELTA updates the base model on newly matured samples, then creates a copy of the base model and adapts the copy on recent pseudo-labeled samples.
The adapted copy is deployed and discarded after serving,
while the base model is retained for subsequent training.
This separation prevents pseudo-label errors from accumulating in the base model and removes the need to correct these updates later. Experiments on real-world datasets demonstrate that DELTA consistently improves LTV prediction performance and support separating temporary adaptation from persistent learning.

\end{abstract}




\maketitle
\section{Introduction}
Recently, live streaming has emerged as an important recommendation scenario on the Web, where users can watch and interact with streamers (i.e., content creators) in real time~\cite{lu2025liveforesighter,gao2023live}. Unlike traditional content recommendation, streamers oftern regularly host live broadcasts, and users often return to watch and interact with the same streamers, fostering long-term user–author (UA) relationships. Previous studies~\cite{zhu2025repeated,guo2026room} have highlighted the importance of modeling such relationships in live-streaming recommendation, as they capture users’ sustained interests in streamers and help promote long-term user engagement and retention beyond immediate interactions.

\begin{figure}[t]
\centering
\includegraphics[width=\columnwidth]{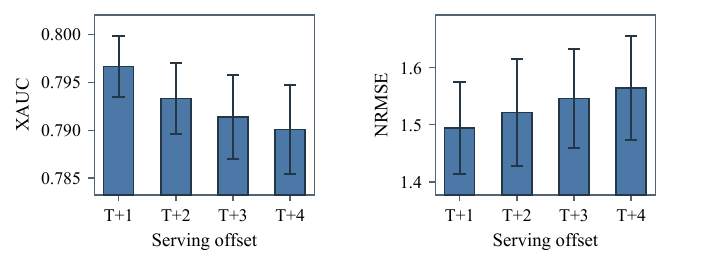}
\caption{Performance degradation as the serving gap increases on KuaiLive-M3. We train a model up to day T and evaluate it on days T + 1 through T + 4 without further updates. As the serving gap increases, XAUC decreases and NRMSE increases, suggesting temporal distribution drift. Bars show means, and error bars indicate standard deviations.}
\label{fig:drift}
\end{figure}

In this work, we focus on predicting the long-term value (LTV) of UA relationships. 
Specifically, for each UA request, we predict the user's cumulative watch time on the same author over a future time window of length $D$.
Such LTV targets can only be obtained after the observation window has closed.
\begin{figure}[t]
\centering
\includegraphics[width=\columnwidth]{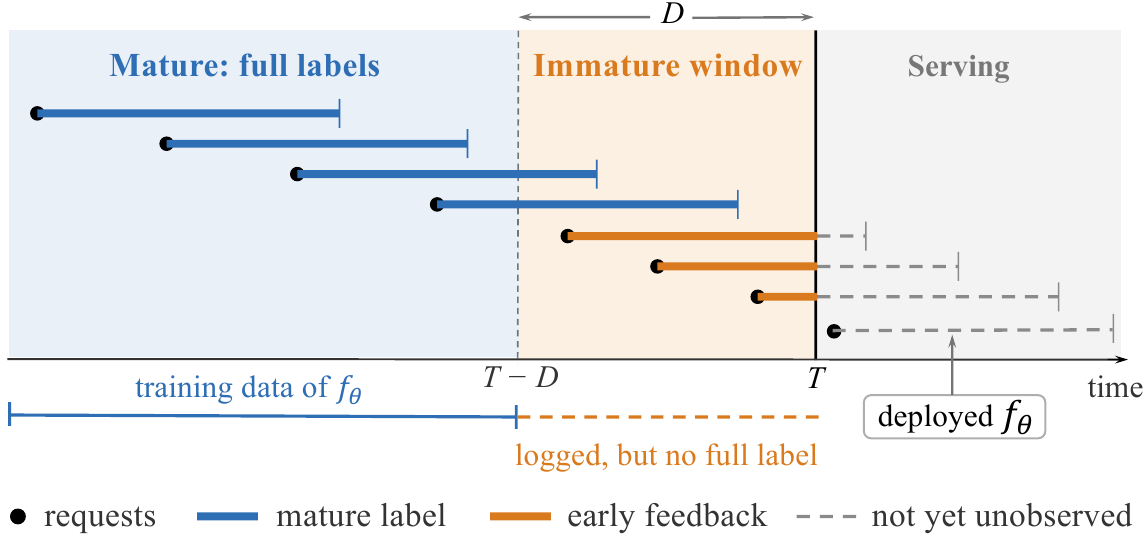}
\caption{Supervision lag in LTV prediction. Each request's full LTV label is available only after $D$ days. At serving time $T$, the model uses current request features but only has been trained on requests before $T-D$. Requests between $T-D$ and $T$ form the immature window: their full labels are unavailable, but the watch time observed so far provides partial feedback.}
\Description{Timeline showing samples with observation windows, mature labels before T-D, and recent samples with partial feedback between T-D and T.}
\label{fig:delay}
\end{figure}
As shown in Figure~\ref{fig:delay}, suppose we are at time $T$, requests between $T-D$ and $T$ do not yet have full LTV labels and are in the immature window. 
Thus, when serving requests at time $T$, the model has only been trained on requests before $T-D$.
In other words, features are fresh, but supervision is stale: the model's input distribution has moved forward to time $T$, while its most recent supervision remains anchored before $T-D$. In an online recommendation system with continuously evolving traffic, this supervision lag may prevent the model from adequately reflecting the recent data distribution. Figure~\ref{fig:drift} empirically illustrates this effect: when a model is trained up to day $T$ and then serves subsequent days without further updates, 
its prediction performance declines over the following days, indicating that data distribution drifts over time.

Although recent samples between $T-D$ and $T$ do not yet have full LTV labels, they already include the watch time observed so far (Figure~\ref{fig:delay}).
This partial feedback contains information about the final LTV, allowing immature samples to be used for training. Some existing methods jointly model LTV targets and short-term targets, using recent immature samples to update a shared model through short-term objectives to keep it fresh ~\cite{ma2018mmoe, tang2020ple, lin2023tree, zhan2022d2q}.
However, the two objectives may produce conflicting gradients, updates on short-term objectives do not necessarily improve LTV prediction~\cite{zou2019reinforcement, cai2023reinforcing, chen2019topk, liu2026discovering}.
Other methods generate pseudo LTV labels for immature samples~\cite{yasui2020fsiw, gu2021defer, yang2021capturing, chen2022asymptotically}.
Yet updates based on these pseudo-labels remain in the model, so estimation errors may accumulate and influence later training.
Even subsequent training on mature true labels and additional correction may not fully eliminate the effects of earlier pseudo-label updates~\cite{li2026delayedgmv,ktena2019addressing, gu2021defer, yang2021capturing}. However, these updates on pseudo-labels need not be retained indefinitely: the corresponding full labels eventually become available and can provide reliable supervision using full true labels. This suggests using early feedback to improve current predictions during the supervision gap, while updating the model retained for future training only with mature true labels.

Motivated by this distinction, we propose  \textbf{DE}ployment-time \textbf{L}earning with \textbf{T}ransient \textbf{A}daptation (\textbf{DELTA}), a framework that uses partial feedback to generate pseudo LTV labels for adaptation on recent samples and decouples this temporary adaptation from persistent learning on mature true labels. Specifically, DELTA trains a privileged pseudo-labeler (PPL) that uses partial feedback as an additional input to generate pseudo LTV labels for recent immature samples, providing supervision for training on these samples.
This feedback serves as privileged information~\cite{vapnik2009lupi} because it is available for historical requests during adaptation but not for new requests at serving time.
At each update, we train the base model on newly matured true labels, obtaining $f^{\mathrm{Base}}$.
We then copy $f^{\mathrm{Base}}$ to initialize a temporary model $f^{\mathrm{Tmp}}$ and adapt it on recent pseudo-labeled samples.
The adapted model is then deployed for serving.
For the next update, we discard this adapted copy and continue training the retained base model $f^{\mathrm{Base}}$ on newly matured true labels.
Since the true labels of recent samples will eventually mature and be used to train the base model, pseudo-labels only need to provide temporary supervision before the true labels mature. Keeping their updates out of the base model therefore prevents estimation errors from accumulating over training and removes the need for later correction. 
Experiments on two real-world datasets demonstrate the effectiveness of DELTA and support decoupling temporary adaptation on pseudo-labeled samples from persistent learning on mature true labels.

Our main contributions are summarized as follows:

\textbf{$\bullet $} We identify supervision lag in UA-level LTV prediction in live-streaming and highlight partial feedback from recent immature samples as a source of timely supervision.

\textbf{$\bullet $} We propose DELTA, a framework that uses partial feedback to generate pseudo LTV labels for adaptation on recent samples and decouples this temporary adaptation from persistent learning on mature true labels.

\textbf{$\bullet $} Experiments on two real-world datasets
demonstrate DELTA's effectiveness in LTV prediction, with
further analyses examining the benefits of temporary adaptation and robustness.

\section{Related Work}
\noindent\textbf{Long-Term Value Prediction.}
Long-term value prediction often involves sparse outcomes and highly skewed value distributions.
Wang et al.~\cite{wang2019deep} introduce a zero-inflated lognormal model to capture both zero-valued outcomes and heavy-tailed positive values, while OptDist~\cite{weng2024optdist} adaptively selects among candidate value distributions.
Another line of work exploits relationships between prediction targets over different time horizons.
Li et al.~\cite{li2022billion} propose ODMN, which uses shorter-horizon LTV prediction tasks to assist longer-horizon prediction by explicitly modeling their ordered dependencies.
GRePO-LTV~\cite{chen2025grepo} combines graph representations with Pareto optimization to address data sparsity and balance prediction objectives across multiple horizons.
Related work also optimizes the lifelong interaction value of UA pairs through reinforcement learning~\cite{li2025rliv}.
These studies advance value distribution modeling, learning across time horizons, and long-term recommendation objectives.
DELTA focuses on using partial feedback from recent immature samples for temporary adaptation while keeping these updates separate from persistent learning on mature labels.

\textbf{Delayed Feedback Modeling.}
Delayed feedback has been extensively studied in conversion prediction, where observed negatives may become positives after additional observation.
ULC~\cite{wang2023ulc} uses an auxiliary model to correct observed negative labels and derives an unbiased label-corrected objective.
For conversion counts and values, Badanidiyuru et al.~\cite{badanidiyuru2021many} exploit observed early outcomes to estimate the remaining value.
MHOL~\cite{gao2022multihead} decomposes outcomes into non-overlapping delay windows and trains a sub-model for each window using its mature labels.
More recently, READER~\cite{li2026delayedgmv} calibrates partial GMV labels into estimated final targets and combines ground-truth alignment with partial label unlearning after the observation window closes.
DELTA also uses partial feedback to estimate long-term targets, but focuses on separating the resulting updates from persistent learning.
Pseudo-labels support adaptation in a temporary serving model, while the base model continues to learn exclusively from mature true labels.
This separation prevents pseudo-label updates from entering the base model and removes the need to correct them later.

\textbf{Live Streaming Recommendation.}
Live streaming recommendation involves dynamic content availability and repeated user--author interactions~\cite{rappaz2021live}. Recent work models room-level collaborative information~\cite{guo2026room} and transfers knowledge from short videos to streamers~\cite{qu2025bridging}. To improve supervision timeliness, Sliver~\cite{liang2024sliver} combines sliding-window sampling with explicit leaving feedback, while MS3M~\cite{zhao2025staytime} uses multiple data streams and segmented objectives to address staytime label truncation. We instead study cumulative UA watch time over a longer observation window, using partial feedback for temporary adaptation while training the retained model only on mature labels.

\section{Problem Formulation}
\label{sec:problem}
\subsection{Daily-Refresh Pipeline}
Since LTV labels require a fixed observation window before they mature, industrial recommender systems typically refresh the model periodically rather than update it continuously from every logged event~\cite{covington2016deep, he2014practical, wang2020practical}. We consider a typical daily-refresh recommendation pipeline, logged requests are collected into daily batches; at the end of a day, the system incrementally updates the model on newly matured samples; then the updated model is deployed for the next serving day; and during serving, request-time features are fetched online while the model remains fixed on that day until the next scheduled update. For LTV prediction, labels mature only after their observation windows close, so the training samples necessarily precede the requests being served.

\subsection{Prediction Task and Label Availability}
We study user--author LTV prediction in live-streaming recommender systems under the daily-refresh pipeline. 
Let $D$ denote the LTV observation window and let $\{d_k\}_{k=1}^{K}$ denote a collection of shorter feedback windows, where $0<d_1<\cdots<d_K<D$. 
DELTA can accommodate different numbers and lengths of short-term label windows through a PPL with one task head for each feedback horizon. Each request $i$ anchors a prediction: given the user, author, and interaction history available at that time, denoted by $x_i$, the model estimates the user's cumulative watch time on the author over the LTV window. 
We denote this LTV label by $y_{D,i}$ and the short-term labels by $y_{d_k,i}$.

\begin{figure}[t]
\centering\small
\includegraphics[width=0.5\textwidth]{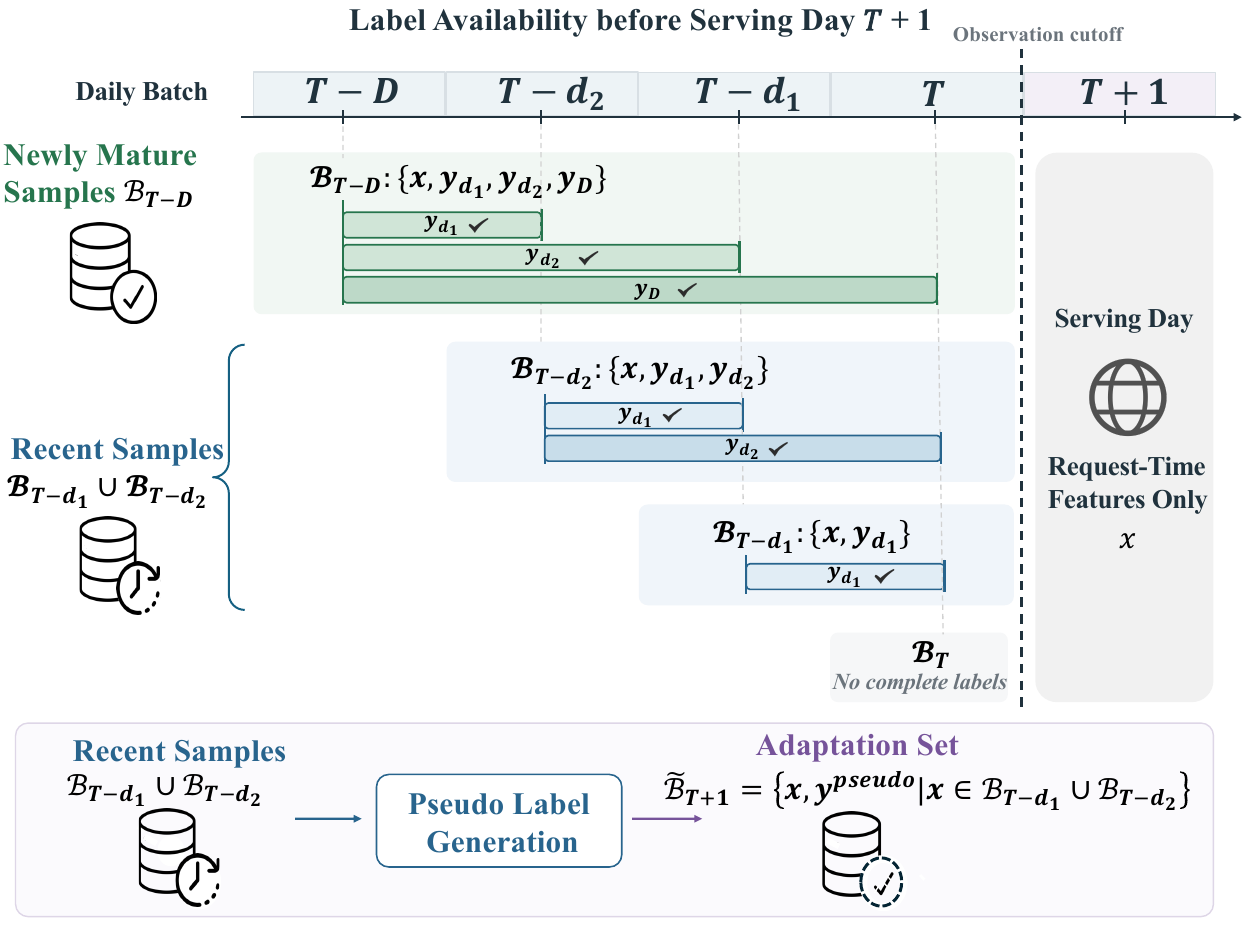}
\caption{An example of label availability and pseudo-label construction before serving day $T+1$, with $D=3$ days, $d_1=1$ day, $d_2=2$ days, and $K=2$.
The newly mature partition $\mathcal B_{T-3}$ contains both short-term labels and the LTV label.
The recent partitions $\mathcal B_{T-2}$ and $\mathcal B_{T-1}$ contain two and one short-term labels, respectively, while $\mathcal B_T$ has no complete label and is excluded.
PPL uses the available partial feedback to generate pseudo LTV labels for the recent partitions, forming the adaptation set $\widetilde{\mathcal B}_{T+1}$.} 
\label{fig:label-availability}
\end{figure}

Let $\mathcal B_t$ denote the requests on day $t$. 
Labels for each daily partition are used only after the corresponding observation window has closed for all its requests. 
Before serving day $T+1$, the newest partition with complete LTV labels is $\mathcal B_{T-D}$, while the newest partition with complete short-term label $y_{d_k}$ is $\mathcal B_{T-d_k}$. 
The partition $\mathcal B_{T-D}$ contains the LTV label and all short-term labels.
For each recent partition $\mathcal B_{T-d_k}$, the short-term labels $\{y_{d_j}\}_{j=1}^{k}$ are available, but the full LTV label remains unavailable.
Figure~\ref{fig:label-availability} illustrates this availability pattern and the construction of pseudo-labels using an example with two short-term feedback windows.
The model is incrementally updated on newly matured samples from $\mathcal B_{T-D}$.
 Our objective is to additionally use the partial feedback from $\{\mathcal B_{T-d_k}\}_{k=1}^{K}$ to improve LTV prediction on day $T+1$. All updates must use only information available before deployment, and serving predictions must depend only on request-time features.

\section{Our Approach: DELTA}
\label{sec:method}
\begin{figure*}[t]
\centering
\includegraphics[width=\textwidth]{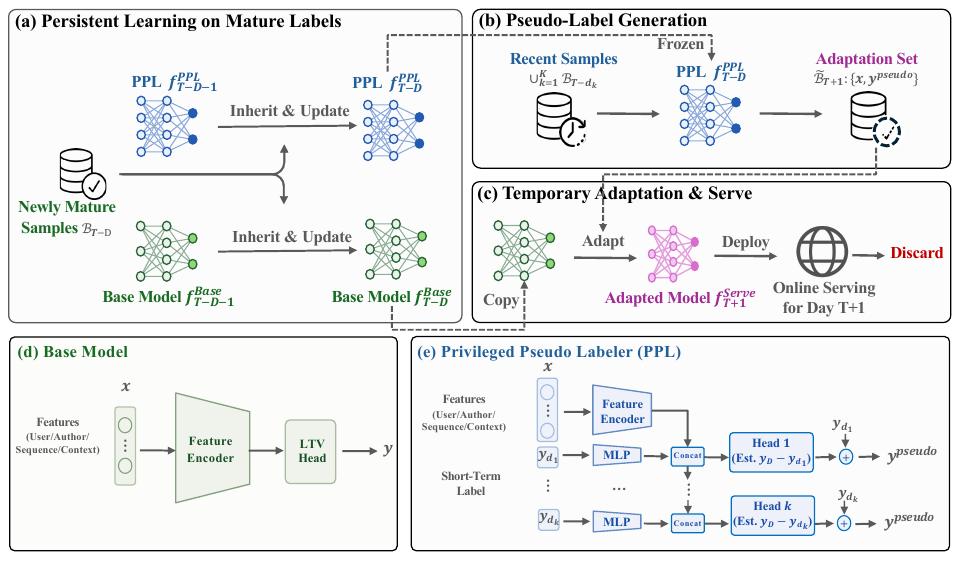}
\caption{Overview of DELTA. (a) Newly mature samples from $\mathcal B_{T-D}$ incrementally update the retained base model and PPL. (b) For samples in $\mathcal B_{T-d_k}$, the frozen PPL applies its $k$-th head to the available short-term labels and generates pseudo LTV labels, forming the adaptation set $\widetilde{\mathcal B}_{T+1}$. (c) A temporary copy of the updated base model is adapted on the adaptation set, then deployed for serving day $T+1$, and discarded afterward. (d) The base model predicts LTV from request-time features only. (e) The PPL shares a feature encoder across horizons and separately encodes each short-term label. Its $k$-th head concatenates the feature representation with the first $k$ label representations to estimate $y_D-y_{d_k}$; adding the observed $y_{d_k}$ produces the pseudo LTV label.}
\label{fig:overview}
\end{figure*}

DELTA uses partial feedback to generate pseudo LTV labels for adaptation on recent data and decouples this temporary adaptation from persistent learning on mature true labels.
As illustrated in Figure~\ref{fig:overview}, each daily update cycle consists of three components: updating the base model and PPL on newly matured samples, generating pseudo-labels for recent immature samples, and adapting a temporary copy of the base model for serving.
We describe these components below.

\subsection{Persistent Learning with Mature Labels}
DELTA maintains a base model that learns only from mature true labels. Before serving day $T+1$, the newest daily batch with complete LTV labels is $\mathcal B_{T-D}$. DELTA incrementally updates the previous base model $f_{T-D-1}^{\mathrm{Base}}$ on this newly matured batch to obtain $f_{T-D}^{\mathrm{Base}}$.
\begin{equation}
 f_{T-D}^{\mathrm{Base}}
 =\operatorname{Update}\!\left(
 f_{T-D-1}^{\mathrm{Base}},\mathcal B_{T-D}
 \right).
 \label{eq:base-update}
\end{equation}
where $\operatorname{Update}$ denotes training with mature true labels.
The base model's subscript indicates the date of the latest mature batch used for training; the update shown above is performed before serving day $T+1$.

Since our LTV target is cumulative watch time over a observation window, the base model adopt weighted logistic regression (WLR), which has been used for watch-time prediction~\cite{covington2016deep,zhan2022d2q}.
Its weighted cross-entropy (WCE) loss accommodates nonnegative continuous labels by using the observed cumulative watch time as the weight of the positive term. For logit $z$ and its sigmoid output $p=\sigma(z)$, given the mature LTV label $y$, the WCE loss is
\begin{equation}
 \begin{aligned}
 \ell_{\mathrm{WCE}}(z,y)
 &=-y\log p-\log(1-p)\\
 &=-yz+(1+y)\log\!\left(1+e^{z}\right).
 \end{aligned}
 \label{eq:wce}
\end{equation}

The base model predicts LTV as $f^{\mathrm{Base}}(x)=\exp(z(x))$.
Its daily update minimizes
\begin{equation}
 \mathcal L_{T+1}^{\mathrm{Base}}
 =\frac{1}{|\mathcal B_{T-D}|}
 \sum_{i\in\mathcal B_{T-D}}
 \ell_{\mathrm{WCE}}\bigl(z(x_i),y_{D,i}\bigr),
 \label{eq:base-loss}
\end{equation}
with model parameters initialized from $f_{T-D-1}^{\mathrm{Base}}$.

The updated base model $f_{T-D}^{\mathrm{Base}}$ is retained for subsequent learning on mature true labels.
Adaptation on recent pseudo-labeled samples takes place only on a separate copy, leaving the base model unaffected.

\subsection{Privileged Pseudo-Labeling}
DELTA uses a privileged pseudo-labeler (PPL) with one task head per feedback horizon. PPL takes request-time features $x$ and the available short-term labels as inputs. These labels serve as privileged information because they are in the immature window and thus are unavailable to the LTV predictor at request time.

PPL contains a shared feature encoder $E$ and a separate label encoder $G_k$ for each short-term label $y_{d_k}$. Their representations are
\begin{equation}
\qquad h_k=G_k(y_{d_k}),\quad k\in\{1,\ldots,K\}.
 \label{eq:ppl-representations}
\end{equation}
The $k$-th task head $H_k$ uses the feature representation and all short-term label representations available by horizon $d_k$:
\begin{equation}
 z_k^{\mathrm{PPL}}
 =H_k\!\left([E(x);h_1;\ldots;h_k]\right),
 \qquad
 \widehat r_k=\exp\!\left(z_k^{\mathrm{PPL}}\right).
 \label{eq:ppl-head}
\end{equation}
Here, $[\,\cdot\,;\,\cdot\,]$ denotes concatenation, and $\widehat r_k$ estimates the remaining value after the $k$-th short-term window. On a mature sample, its target is
\begin{equation}
 r_k=y_D-y_{d_k},\qquad k\in\{1,\ldots,K\}.
 \label{eq:ppl-residual}
\end{equation}

Before serving day $T+1$, PPL is updated on $\mathcal B_{T-D}$, where all short-term labels and LTV label are fully observed. The shared feature encoder, label encoders, and task heads are trained jointly by minimizing 

\begin{equation}
 \mathcal L_{T+1}^{\mathrm{PPL}}
 =
 \frac{1}{K|\mathcal B_{T-D}|}
 \sum_{i\in\mathcal B_{T-D}}\sum_{k=1}^{K}
 \ell_{\mathrm{WCE}}\!\left(
 z_{k,i}^{\mathrm{PPL}},y_{D,i}-y_{d_k,i}
 \right).
 \label{eq:ppl-loss}
\end{equation}


The PPL continues learning from newly mature samples across days and remains fixed while generating pseudo-labels. For a recent sample with feature $x$ from $\mathcal B_{T-d_k}$, the first $k$ short-term labels $y_{d_1},...,y_{d_k}$ are available. DELTA applies the $k$-th task head to the sample and constructs its pseudo LTV label as
\begin{equation}
y^{pseudo}(x)
 =
 y_{d_k}
 +
 \exp\!\left(
 H_k([E(x);G_1(y_{d_1});\ldots;G_k(y_{d_k})])
 \right).
 \label{eq:pseudo-label}
\end{equation}

These pseudo-labeled samples form the adaptation set $\widetilde{\mathcal B}_{T+1}$,
\begin{equation}
 \widetilde{\mathcal B}_{T+1} = \{x,y^{pseudo}(x)| x\in \bigcup_{k=1}^K \mathcal B_{T-d_k}\},\
\end{equation}

Although a pseudo-label may differ from an individual sample's true LTV, it can still provide suitable supervision if it captures the average outcome given the available information. Let $S_k=(Y_{d_1},\ldots,Y_{d_k})$ denote the short-term feedback available to the $k$-th head. An ideal head predicts the conditional mean of the remaining value, yielding
\begin{equation}
 \begin{aligned}
 q_k(X,S_k)
 &=
 Y_{d_k}+
 \mathbb E[Y_D-Y_{d_k}\mid X,S_k]\\
 &=
 \mathbb E[Y_D\mid X,S_k].
 \end{aligned}
 \label{eq:ppl-conditional-mean}
\end{equation}
The second equality holds because $Y_{d_k}$ is included in $S_k$. For a fixed prediction, WCE depends linearly on the label as shown in Eq.~\eqref{eq:wce}. Thus, using the ideal pseudo-label $q_k(X,S_k)$ instead of the true LTV label does not change the expected WCE loss. For any request-time logit predictor $z$, the tower property gives
\begin{equation}
 \begin{aligned}
 \mathbb E\!\left[
 \ell_{\mathrm{WCE}}(z(X),q_k(X,S_k))
 \right]=
 \mathbb E\!\left[
 \ell_{\mathrm{WCE}}(z(X),Y_D)
 \right].
 \end{aligned}
 \label{eq:tower}
\end{equation}

Thus, if each task head of PPL recovers this conditional mean, its pseudo-label give the same expected WCE loss as true labels.
This provides a rationale for using them to supervise adaptation on recent samples. In practice, estimation error and distribution shifts may affect their quality, which we evaluate in Section~\ref{sec:exp-teacher}. The next section describes how DELTA uses these labels for temporary adaptation while keeping their updates separate from base-model training.

\subsection{Temporary adaptation and Serving}
\label{sec:separation}
Before deployment, DELTA copies the updated base model $f_{T-D}^{\mathrm{Base}}$ to initialize a temporary model $f_{T+1}^{\mathrm{Tmp}}$. This copy is adapted on the pseudo-labeled recent samples by minimizing
\begin{equation}
 \mathcal L_{T+1}^{\mathrm{Adapt}}(f)
 =\frac{1}{|\widetilde{\mathcal B}_{T+1}|}
 \sum_{(x_i,y_i^{\mathrm{pseudo}})\in\widetilde{\mathcal B}_{T+1}}
 \ell_{\mathrm{WCE}}\bigl(\log f(x_i),y_i^{\mathrm{pseudo}}\bigr).
 \label{eq:adapt}
\end{equation}
After adaptation, the model is denoted by $f_{T+1}^{\mathrm{Serve}}$ and deployed for day $T+1$. It remains fixed during serving and uses only request-time features. The procedure is summarized as
\begin{equation}
 f_{T-D}^{\mathrm{Base}}
 \xrightarrow{\quad\mathrm{copy}\quad}
 f_{T+1}^{\mathrm{Tmp}}
 \xrightarrow{\mathrm{adapt\ on}\ \widetilde{\mathcal B}_{T+1}}
 f_{T+1}^{\mathrm{Serve}}.
 \label{eq:serving-pipeline}
\end{equation}

After serving day $T+1$, $f_{T+1}^{\mathrm{Serve}}$ is discarded. At the next daily update, the retained base model learns from the newly matured partition $\mathcal B_{T-D+1}$:
\begin{equation}
 f_{T-D+1}^{\mathrm{Base}}
 =\operatorname{Update}\bigl(f_{T-D}^{\mathrm{Base}},\mathcal B_{T-D+1}\bigr),
 \label{eq:base}
\end{equation}
The updated base model is then copied for a new round of adaptation before serving day $T+2$.

As time progresses and new labels mature, DELTA continues this daily update cycle, using pseudo-labels for current serving while retaining only mature-label updates in the base model. For a fixed stream of logged data, the base model follows the same training trajectory as learning only from mature labels. Each recent sample can support adaptation through its pseudo-label and later enter base-model training with its true label, without any earlier pseudo-label update to undo.

\section{Experiments}
\label{sec:exp}
We evaluate DELTA through the following research questions:

\noindent\textbf{$\bullet $} \textbf{RQ1:} How does DELTA perform compared with other baselines?

\noindent\textbf{$\bullet $} \textbf{RQ2:} How accurate are PPL's pseudo-labels?

\noindent\textbf{$\bullet $} \textbf{RQ3:} How does temporary adaptation compare with persistent adaptation?

\noindent\textbf{$\bullet$} \textbf{RQ4:} How robust is DELTA to pseudo-label quality?

\noindent\textbf{$\bullet$} \textbf{RQ5:} How much additional overhead does DELTA introduce?

\subsection{Experimental Setup}
Throughout our experiments, we set the LTV observation window to $D=3$ days and use $K=2$ short-term feedback windows, with $d_1=1$ day and $d_2=2$ days.
Thus, $y_D$ denotes cumulative watch time over three days, while $y_{d_1}$ and $y_{d_2}$ denote cumulative watch time over one and two days, respectively. More experimental details can be found in Appendix~\ref{sec:implementation}.
\subsubsection{Datasets}
We conduct experiments on two public datasets: KuaiLive-M3~\cite{guo2026m3}, a large-scale live-streaming recommendation dataset, and KuaiRand-1k~\cite{gao2022kuairand}, a short-video recommendation dataset used to examine whether DELTA extends beyond live streaming. Table~\ref{tab:datasets} summarizes the processed datasets and evaluation periods.

\textbf{KuaiLive-M3}. KuaiLive-M3 is a large-scale public dataset containing approximately 35 million user–streamer interactions. It records users’ viewing behaviors in live streaming, allowing us to construct cumulative watch-time targets for long-term user–author engagement. We use this dataset as our primary benchmark for evaluating LTV prediction under delayed supervision.

\textbf{KuaiRand-1k}. KuaiRand-1k is a public recommendation dataset containing short-video interaction logs from approximately 1,000 users. These logs include interaction timestamps, watch time, and other user feedback, supporting the construction of temporally ordered samples and cumulative watch-time targets. We use this dataset to examine whether DELTA also benefits LTV prediction beyond live streaming, where content consumption and interaction patterns differ.

\begin{table}[t]
\caption{Statistics of the datasets in our experiments.}
\label{tab:datasets}
\centering\small
\setlength{\tabcolsep}{7pt}
\begin{tabular}{lrr}
\toprule
Statistic & KuaiRand-1k & KuaiLive-M3\\
\midrule
Users & 999 & 21,937\\
Authors & 959,122 & 554,731\\
Total samples & 7,559,706 & 13,079,435\\
Total days & 27 & 26\\
Serving days & 11 & 11\\
\bottomrule
\end{tabular}
\end{table}

\subsubsection{Baselines}
We compare DELTA with the following baselines:
\begin{itemize}
    \item \textbf{Daily Base} is updated exclusively on samples with mature LTV labels. Before serving day $T+1$, it incrementally learns from $\mathcal B_{T-D}$ without using recent immature samples.
    
    \item \textbf{Synchronous MTL} uses a shared encoder with separate heads for the two short-term targets and LTV target. It waits until all three labels are mature and updates the three tasks jointly on the same partition $\mathcal B_{T-D}$.

    \item \textbf{Asynchronous MTL} uses the same architecture as Synchronous MTL but updates each task on its newest eligible partition, without waiting for the longer-horizon labels to mature. Before serving day $T+1$, the LTV task is updated on $\mathcal B_{T-D}$, while each short-term task $k$ is updated on $\mathcal B_{T-d_k}$. Each task updates the shared encoder and its own head.

    \item \textbf{Recursive}~\cite{badanidiyuru2021many} recursively combines observed feedback with auxiliary predictions to construct pseudo-labels for recent samples. The main predictor retains its pseudo-label updates and does not revisit these samples for true-label training when their final labels mature.

    \item \textbf{GRePO-LTV}~\cite{chen2025grepo} combines graph representations with Pareto optimization to jointly predict value over multiple horizons. We adapt it to predict the two short-term labels and the LTV label under the same label-availability constraints.
    
    \item \textbf{READER}~\cite{li2026delayedgmv} uses the same PPL-generated pseudo-labels as DELTA but retains their updates across days. Once true labels mature, it applies the correction \(1.1\,\mathrm{GRA}-0.05\,\mathrm{PLU}\) to incorporate ground-truth alignment and partially unlearn earlier pseudo-label supervision.

\end{itemize}
All methods follow the chronological label-availability schedule in Figure~\ref{fig:label-availability} and are evaluated on the same serving-day samples.

\subsubsection{Metrics}
 LTV prediction requires accurate value estimates and reliable rankings, since recommender systems use predicted LTV to prioritize candidates with greater LTV. We therefore report XAUC, Spearman correlation, and normalized Gini (NormGini) to evaluate ranking quality, together with RMSE and mean-normalized RMSE (NRMSE) to evaluate prediction accuracy. Higher ranking metrics and lower error metrics indicate better performance. Metrics are computed within windows of 512 examples and averaged equally over valid windows within each day. Daily metrics are then averaged equally across serving days.

\subsection{RQ1: Overall Prediction Performance}
\label{sec:exp-performance}

\begin{table*}[t]
\caption{Prediction performance of deployable methods. Higher XAUC, Spearman, and Gini are better; lower RMSE and NRMSE are better. The best and second-best results within each dataset are marked in bold and underlined, respectively.}
\label{tab:main}
\centering\small
\setlength{\tabcolsep}{7pt}
\begin{tabular}{llrrrrr}
\toprule
Dataset & Method & XAUC$\uparrow$ & Spearman$\uparrow$ & Gini$\uparrow$ & RMSE$\downarrow$ & NRMSE$\downarrow$ \\
\midrule
\multirow{7}{*}{KuaiLive-M3}
& Daily Base & 0.7901 & 0.7540 & 0.8437 & 5839.85 & 1.5519 \\
& Synchronous MTL & 0.7890 & 0.7515 & 0.8424 & \underline{5596.61} & \underline{1.4875} \\
& Asynchronous MTL & 0.7861 & 0.7461 & 0.8434 & 5897.97 & 1.5672 \\
& Recursive & 0.7927 & 0.7581 & \underline{0.8499} & 6105.27 & 1.6191 \\
& GRePO-LTV & 0.7911 & 0.7559 & 0.8356 & 5852.15 & 1.5530 \\
& READER & \underline{0.7933} & \underline{0.7599} & 0.8498 & 5947.03 & 1.5811 \\
& DELTA & \textbf{0.7939} & \textbf{0.7610} & \textbf{0.8503} & \textbf{5194.89} & \textbf{1.3790} \\
\midrule
\multirow{7}{*}{KuaiRand-1k}
& Daily Base & 0.7615 & 0.3930 & 0.7262 & 43.36 & 5.9658 \\
& Synchronous MTL & 0.7628 & 0.3948 & 0.7278 & 43.62 & 5.9950 \\
& Asynchronous MTL & 0.7591 & 0.3893 & 0.7277 & 45.75 & 6.2362 \\
& Recursive & 0.7635 & 0.3957 & 0.7131 & 43.52 & 6.0331 \\
& GRePO-LTV & 0.7633 & 0.3954 & \textbf{0.7338} & 46.25 & 6.3495 \\
& READER & \underline{0.7669} & \underline{0.4007} & 0.7305 & \underline{42.84} & \underline{5.9178} \\
& DELTA & \textbf{0.7671} & \textbf{0.4011} & \underline{0.7310} & \textbf{42.83} & \textbf{5.8990} \\
\bottomrule
\end{tabular}
\end{table*}

The experimental results are summarized in Table~\ref{tab:main}, from which we draw the following observations.
\begin{enumerate}
    \item \textbf{DELTA achieves the strongest overall performance.}
    Compared with all baselines, DELTA delivers the strongest overall performance across both datasets. These results demonstrate its effectiveness in LTV prediction under delayed supervision.

    \item \textbf{DELTA outperforms methods that jointly optimize LTV and short-term objectives.}
Synchronous MTL jointly learns LTV and short-term objectives on mature samples, whereas Asynchronous MTL additionally updates the shared representation using short-term labels from recent immature samples.
GRePO-LTV uses Pareto optimization to balance objectives across prediction horizons.
DELTA achieves better overall performance than these baselines, including Asynchronous MTL, which also benefits from recent supervision.
These results support using partial feedback to generate pseudo LTV labels for adaptation directly on the LTV objective.

    \item \textbf{Comparisons with pseudo-label baselines support separating temporary adaptation from persistent learning.}
    Recursive trains its main predictor with pseudo-labels without revisiting the same samples for true-label training once they mature. READER retains pseudo-label updates, then incorporates mature true labels and corrects the effects of earlier pseudo-label supervision. DELTA outperforms both methods on all reported metrics across both datasets. These results suggest that confining pseudo-label updates to the current serving period is a useful design choice. RQ3 further examines this choice through a direct comparison with persistent adaptation.
\end{enumerate}

\subsection{RQ2: Quality of Pseudo-Labels}
\label{sec:exp-teacher}

We test whether PPL-generated pseudo-labels are more accurate than base-model predictions on recent immature samples.
For each serving day $T+1$, the $k$-th head of PPL generates pseudo-labels for $\mathcal B_{T-d_k}$.
The base model predicts LTV on the same samples.
All models are trained on mature data through $\mathcal B_{T-D}$.
Following~\cite{liu2026discovering}, we compute evaluation metrics by comparing these predictions with the true LTV labels that become available later.

\begin{table*}[t]
\caption{Performance of base-model predictions and PPL-generated pseudo-labels. For each \(k\), pseudo-labels from the \(k\)-th PPL head are evaluated on \(\mathcal B_{T-d_k}\). Base and PPL rows report absolute metrics; italic rows report relative changes from Base to PPL. Positive changes in ranking metrics and negative changes in error metrics indicate improvement.}
\label{tab:teachers}
\centering\small
\setlength{\tabcolsep}{7pt}
\begin{tabular}{lclrrrrr}
\toprule
Dataset & Partition & Model & XAUC\(\uparrow\) & Spearman\(\uparrow\) & Gini\(\uparrow\) & RMSE\(\downarrow\) & NRMSE\(\downarrow\)\\
\midrule
\multirow{6}{*}{KuaiLive-M3}
& \multirow{3}{*}{\(\mathcal B_{T-d_1}\)} & Base & 0.7933 & 0.7599 & 0.8536 & 5637.50 & 1.5085\\
& & PPL & 0.8815 & 0.9034 & 0.9427 & 3230.58 & 0.8647\\
& & \textit{Improvement} & \textit{+11.1\%} & \textit{+18.9\%} & \textit{+10.4\%} & \textit{-42.7\%} & \textit{-42.7\%}\\
\addlinespace[3pt]
& \multirow{3}{*}{\(\mathcal B_{T-d_2}\)} & Base & 0.7966 & 0.7656 & 0.8626 & 5488.78 & 1.4804\\
& & PPL & 0.9324 & 0.9553 & 0.9809 & 1842.73 & 0.4987\\
& & \textit{Improvement} & \textit{+17.0\%} & \textit{+24.8\%} & \textit{+13.7\%} & \textit{-66.4\%} & \textit{-66.3\%}\\
\midrule
\multirow{6}{*}{KuaiRand-1k}
& \multirow{3}{*}{\(\mathcal B_{T-d_1}\)} & Base & 0.7642 & 0.3990 & 0.7276 & 41.50 & 5.5883\\
& & PPL & 0.8541 & 0.5243 & 0.8571 & 27.55 & 3.6691\\
& & \textit{Improvement} & \textit{+11.8\%} & \textit{+31.4\%} & \textit{+17.8\%} & \textit{-33.6\%} & \textit{-34.3\%}\\
\addlinespace[3pt]
& \multirow{3}{*}{\(\mathcal B_{T-d_2}\)} & Base & 0.7668 & 0.4039 & 0.7330 & 40.98 & 5.3868\\
& & PPL & 0.9248 & 0.6150 & 0.9379 & 16.51 & 2.1727\\
& & \textit{Improvement} & \textit{+20.6\%} & \textit{+52.3\%} & \textit{+28.0\%} & \textit{-59.7\%} & \textit{-59.7\%}\\
\bottomrule
\end{tabular}
\end{table*}

As shown in Table~\ref{tab:teachers}, pseudo-labels from the PPL's two horizon-specific heads outperform base-model predictions on recent samples across both datasets. They reduce RMSE by 33.6\%--66.4\% and improve all other reported metrics.

The PPL and base model are trained on the same mature history, but the PPL additionally uses observed short-term feedback to estimate the eventual LTV. Together with the conditional-mean analysis in Eq.~\eqref{eq:tower}, these results provide theoretical and empirical justification for using PPL-generated pseudo-labels to supervise adaptation on recent immature samples.

\subsection{RQ3: Temporary versus Persistent Adaptation}
\label{sec:exp-lifetime}

In this section, we test whether discarding pseudo-label updates after serving helps prevent errors from accumulating in later training.
We compare DELTA with a persistent adaptation variant on KuaiLive-M3.
The two variants use the same model architecture, pseudo-labels, and training settings, differing only in whether the parameters updated during adaptation are carried forward to the next update.
The persistent variant, called DELTA (Persistent), retains the adapted serving model and continues training it on newly matured true labels and recent pseudo-labels in the next update.
DELTA discards the adapted copy after serving and starts the next update from the retained base model, which learns only from mature true labels.

\begin{figure}[t]
\centering
\includegraphics[width=\columnwidth]{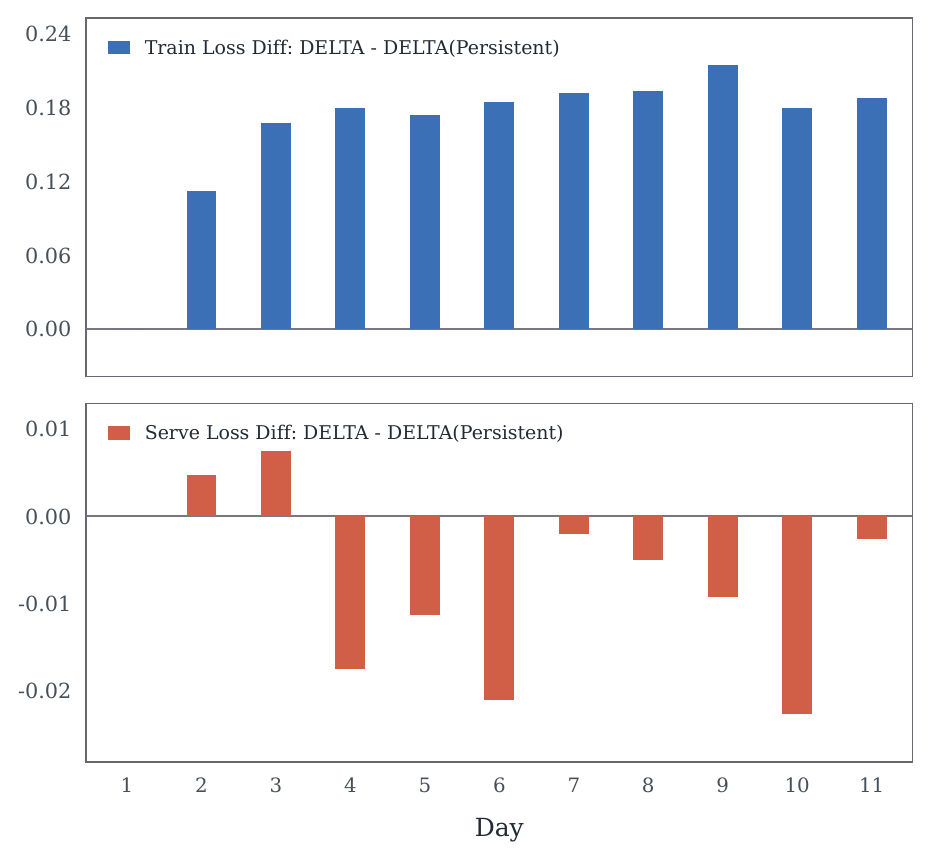}
\caption{Training and serving losses of temporary and persistent adaptation on KuaiLive-M3. Persistent adaptation achieves lower training loss but higher serving loss in the
later evaluation days.}
\label{fig:adaptation-loss}
\end{figure}

Table~\ref{tab:lifetime} shows that DELTA outperforms the persistent variant on all five metrics. 
To examine how this difference develops over time, Figure~\ref{fig:adaptation-loss} compares their training loss on newly matured samples and serving loss on current-day requests. 
The persistent variant achieves lower training loss, and this gap generally widens over time.
However, its increasingly better fit to training samples does not translate into a consistent serving advantage; instead, its serving loss is higher than DELTA's in the later evaluation days.
These results suggest that retaining pseudo-label updates improves the fit to training samples without improving predictions on new requests.

Two factors may explain this difference.
First, pseudo-labels contain estimation errors, and retaining pseudo-label updates can carry estimation errors into later training, even after true labels become available.
Second, the persistent model has already trained on newly matured samples using their earlier pseudo-labels.
When these samples enter mature-label training, the model parameters already reflect feedback from the same requests.
At serving time, however, predictions for new requests must be made before their feedback becomes available.
This creates a mismatch between training and serving, similar to the one caused by earlier feedback updates in multi-window staytime prediction~\cite{liu2026discovering}.
Such prior use of feedback can make mature samples easier to fit without necessarily improving predictions on new requests.

DELTA's base model learns from each sample only after its true label matures.
Earlier pseudo-label updates remain in temporary serving copies and are discarded afterward.
DELTA's lower serving loss and better prediction results support this design.

\begin{table}[t]
\caption{Temporary versus persistent adaptation on KuaiLive-M3. Higher ranking metrics and lower error metrics are better.}
\label{tab:lifetime}
\centering\small
\setlength{\tabcolsep}{8pt}
\begin{tabular}{lrr}
\toprule
Metric & DELTA & DELTA (Persistent)\\
\midrule
XAUC$\uparrow$ & \textbf{0.7939} & 0.7936\\
Spearman$\uparrow$ & \textbf{0.7610} & 0.7605\\
Gini$\uparrow$ & \textbf{0.8503} & 0.8491\\
RMSE$\downarrow$ & \textbf{5194.89} & 5252.70\\
NRMSE$\downarrow$ & \textbf{1.3790} & 1.3949\\
\bottomrule
\end{tabular}
\end{table}

\subsection{RQ4: Robustness to Pseudo-Label Quality}
\label{sec:exp-robustness}
We compare DELTA with DELTA (Persistent) on KuaiLive-M3
under controlled label perturbations.
For this offline diagnostic, we replace PPL predictions
with a mixture of true labels and labels randomly shuffled
within each daily partition:
\begin{equation}
 \widetilde y_{D,i}^{(\alpha)}
 =
 (1-\alpha)y_{D,i}
 +
 \alpha y_{D,i}^{\mathrm{shuffle}},
 \label{eq:controlled-pseudo-noise}
\end{equation}
where $y_{D,i}^{\mathrm{shuffle}}$ is the label assigned to
sample $i$ after shuffling.
The parameter $\alpha$ controls the perturbation strength,
with $\alpha=0$ recovering true labels and $\alpha=1$
using fully shuffled labels.
Both methods use identical perturbed labels under matched
training settings.

\begin{figure}[t]
    \centering
\includegraphics[width=0.9\columnwidth]{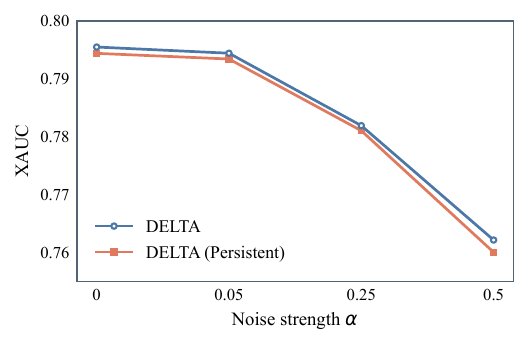}
    \caption{XAUC of DELTA and DELTA (Persistent) under controlled pseudo-label noise on KuaiLive-M3, evaluated at $\alpha\in\{0,0.05,0.25,0.5\}$.}
    \label{fig:pseudo-quality-robustness}
\end{figure}

As shown in Figure~\ref{fig:pseudo-quality-robustness},
DELTA achieves higher XAUC than the persistent
variant at every evaluated noise level. This advantage persists even at $\alpha=0$, where adaptation
uses true labels. This is consistent with the explanation in RQ3 that repeated updates on the same requests may improve training fit
without improving generalization. The gap is larger under strong perturbation ($\alpha=0.5$),
suggesting that keeping adaptation updates
temporary can improve robustness to unreliable pseudo-labels.
Nevertheless, both methods deteriorate as
noise increases, highlighting the importance of informative supervision.

\subsection{RQ5: Additional Overhead}
\label{sec:exp-cost}
DELTA introduces three sources of additional overhead compared with daily base-model training. First, it maintains a PPL with a shared feature encoder, $K$ short-term label encoders, and $K$ task heads, which are jointly updated on newly mature samples. Second, it processes the recent partitions $\{\mathcal B_{T-d_k}\}_{k=1}^{K}$ and applies the corresponding task head to generate pseudo-labels for each partition. Third, it creates a temporary copy of the base model and trains it on these pseudo-labeled samples before deployment.

All additional computation and temporary model storage are confined to the pre-deployment stage. During serving, DELTA uses a single LTV predictor with the same architecture and request-time inputs as the base model. Neither the PPL nor adaptation is involved in online prediction, so DELTA introduces no additional serving-time computation or model-memory requirement.

\section{Discussion}
DELTA is motivated by a distinction between the immediate value of early feedback and the need to retain its updates over time. Before full labels mature, partial feedback provides timely information that mature-only training cannot yet exploit. Once the corresponding full labels become available, these requests can contribute to persistent learning through reliable supervision. This suggests that pseudo-labels can serve a temporary role: improving predictions during the supervision gap without requiring their updates to become part of the retained model. DELTA trains its base model only on mature true labels,
while using temporary adaptation to incorporate recent feedback.
This allows the serving model to respond to recent interactions
without carrying pseudo-label errors or the effects of repeated
training into subsequent base model updates.
The separation also has limits. Adaptation on pseudo-labels
does not guarantee better serving performance, and discarding
the updates may also lose useful information learned during
adaptation. A promising direction for future work is to let the adaptation strategy evolve over time, using past adaptation experience and serving outcomes, evaluated once true labels mature, to guide how to do beneficial adaptation in future periods.

\section{Conclusion}
We propose DELTA to address supervision lag in user--author LTV prediction for live-streaming recommendation. DELTA uses partial feedback to generate pseudo LTV labels for adaptation on recent samples and decouples this temporary adaptation from persistent learning on mature true labels. Experiments on two real-world datasets show improved overall performance, and comparisons with persistent adaptation support the value of separating the adaptation and persistent learning on mature samples. These findings suggest that benefiting from early feedback does not necessarily require retaining its updates indefinitely. When complete labels eventually arrive, temporary adaptation offers a practical way to use timely but imperfect supervision for current predictions while reserving persistent learning for mature labels.

\bibliographystyle{ACM-Reference-Format}
\bibliography{sref}

\appendix
\section{Implementation Details}
\label{sec:implementation}
The base model truncates the user and user--author sequences to 1,000 and 50 events, respectively. Identifier and categorical embeddings have width 8, while sequence and attention representations have width 32. Its encoder has hidden width 128 and output width 64, followed by a width-64 prediction tower with dropout 0.1. The output is the exponential of a logit capped at 11 for numerical stability.

The PPL uses a shared width-64 feature encoder and a separate two-layer width-64 label encoder for each short-term label. The $k$-th head concatenates the feature representation with the first $k$ label representations, then applies a width-64 fusion layer to predict the residual logit. The base model is pretrained for 10 epochs, while the PPL uses 15 epochs on the same mature-data schedule. KuaiRand-1k uses 12 pretraining days and KuaiLive-M3 uses 11; both use 11 daily evaluation cycles.

The label cap is 6,000 seconds on KuaiRand-1k and 40,000 seconds on KuaiLive-M3. The base model and PPL use a gradient-clipping threshold of 1,000; adaptation uses 30 and 100, respectively. Training and evaluation batch sizes are 2,048 and 512.

\end{document}